\documentclass{ecai}
\usepackage{graphicx}
\usepackage{latexsym}
\usepackage[disable]{todonotes}
\usepackage{amssymb}

\usepackage{algorithm}
\usepackage[noend]{algpseudocode}
\makeatletter
\renewcommand{\ALG@beginalgorithmic}{\footnotesize}
\makeatother
\algrenewcommand\algorithmicrequire{\textbf{Input:}}
\algrenewcommand\algorithmicensure{\textbf{Output:}}

\usepackage{tikz}
\usepackage{pgfplots}
\usepackage{subcaption}
\pgfplotsset{compat=newest}

\pgfplotsset{every x tick label/.append style={font=\scriptsize, yshift=0.4ex}}
\pgfplotsset{every y tick label/.append style={font=\scriptsize, xshift=0.4ex}}
\pgfplotsset{every axis/.append style={
                    axis line style={<->}, 
                    xlabel={$x$},          
                    ylabel={$y$},          
                    label style={font=\scriptsize},
                    tick label style={font=\scriptsize}  
                    }}

\ecaisubmission   

\begin{document}

\begin{frontmatter}

\title{Learning Environment Models with \\ Continuous Stochastic Dynamics}

\author[A,B]{\fnms{Martin}~\snm{Tappler}\thanks{Corresponding Author. Email: martin.tappler@ist.graz.at.}}
\author[B,A]{\fnms{Edi}~\snm{Mu\v{s}kardin}}
\author[A]{\fnms{Bernhard K.}~\snm{Aichernig}} 
\author[C]{\fnms{Bettina}~\snm{K\"onighofer}} 

\address[A]{Institute of Software Technology, Graz University of Technology, Graz, Austria}
\address[B]{
TU Graz - SAL DES Lab, Silicon Austria Labs, Graz, Austria}

\address[C]{
Institute of Applied Information Processing and Communications, Graz University of Technology, Graz, Austria }

\begin{abstract}
 Solving control tasks in complex environments automatically through learning offers great potential. While contemporary techniques from deep reinforcement learning (DRL) provide effective solutions, their decision-making is not transparent. We aim to provide insights into the decisions faced by the agent by learning an automaton model of environmental behavior under the control of an agent. However, for most control problems, automata learning is not scalable enough to learn a useful model. In this work, we raise the capabilities of automata learning such that it is possible to learn models for environments that have \emph{complex} and \emph{continuous dynamics}.

The core of the scalability of our method lies in the computation of an abstract state-space representation, by applying \emph{dimensionality reduction} and \emph{clustering} on the observed environmental state space. The stochastic transitions are learned via passive automata learning from observed interactions of the agent and the environment. In an iterative model-based RL process, we sample additional trajectories to learn an accurate environment model in the form of a discrete-state Markov decision process (MDP). We apply our automata learning framework on popular RL benchmarking environments in the OpenAI Gym, including LunarLander, CartPole, Mountain Car, and Acrobot. Our results show that the learned models are so precise that they enable the computation of policies solving the respective control tasks. Yet the models are more concise and more general than neural-network-based policies and by using MDPs we benefit from a wealth of tools available for analyzing them. When solving the task of LunarLander, the learned model even achieved similar or higher rewards than deep RL policies learned with stable-baselines3.

\end{abstract}

\end{frontmatter}

\newcommand{\ALERGIA}{\textsc{Alergia}}
\newcommand{\IOALERGIA}{\textsc{IOAlergia}}
\newcommand{\AALpy}{\textsc{AALpy}}

\newcommand{\Dist}{\mathit{Dist}}
\newcommand{\lab}{L}

\newcommand{\mdp}{\mathcal{M}}
\newcommand{\trans}{\mathcal{P}}
\newcommand{\states}{\mathcal{S}}
\newcommand{\Act}{\mathcal{A}}

\newcommand{\policy}{\sigma}
\newcommand{\policies}{\Sigma}

\section{Introduction}

In an ideal world, an interpretable, correct, and compact model of any complex system (operating in a complex environment) should be available before the system is deployed. Having such a model allows to perform a comprehensive analysis whether the system adheres to critical properties. 
However,  concise models that allow a comprehensive analysis of the system are rarely available. 

\emph{Automata learning}, often called \emph{model learning}, is a 
widely used technique to infer a finite-state model from a given black-box system just by observing its behavior~\cite{DBLP:conf/uss/RuiterP15,DBLP:conf/cav/Fiterau-Brostean16,DBLP:conf/icst/TapplerAB17}.
The inferred model can then be used to detect undesired or unsafe system behavior.

Automatically generating controllers for environments with \emph{continuous state space} and \emph{complex stochastic dynamics}
through machine learning has great potential, since analytical 
solutions require immense human effort. 
Model-free deep reinforcement
learning (DRL) especially has proven successful in solving complex sequential decision making problems in high-dimensional, probabilistic environments. 
However, the decision making of deep learning systems is highly opaque and the lack of having explainable models limits their acceptance in promising application areas like autonomous mobility, medicine, or finance.

Thus, it is of particular importance to have methods and tools available that automatically
learn environmental models under the control of an agent. 
However, the high-dimensionality of the
observed environmental states and the 
complex environment's dynamics 
renders a direct application of automata learning infeasible. 

\noindent\textbf{CASTLE - Clustering-based Activated pasSive automTa LEarning}. CASTLE is a novel approach for learning environmental models which is especially designed to cope with \emph{environments with complex dynamics and continuous state spaces}.

Having such a model available, makes it possible to compute probabilities of how likely it is that executing an action will result in successfully completing the task under consideration. Via the application of probabilistic model checking~\cite{Baier2018}, these probabilities can be computed fully automatically, for any state and action in the learned model. This  
allows us to analyze the decision making of an agent and relate the agent’s policy to any other possible policy. 

\noindent\textbf{Problem statement.}
Our goal is to learn a  finite-state MDP representing the environment
under the control of an agent solving an episodic task.
We aim to learn MDPs that are sufficiently accurate to compute effective decision-making policies.
It should be emphasized that by "learning an MDP", we mean learning its complete structure, including the states and transitions, not just its probabilities.

\begin{figure*}[t]
    \centering
    \includegraphics[width=0.83\textwidth] {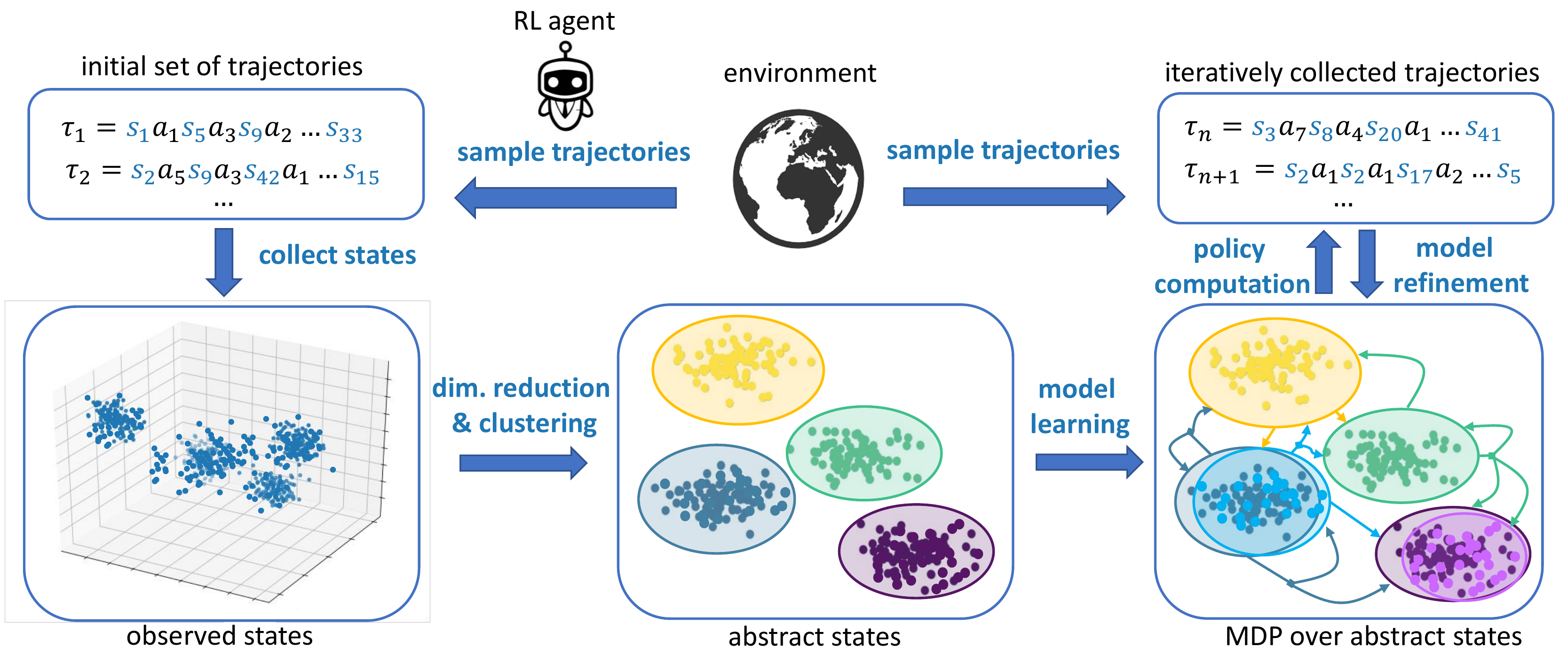}
    \caption{Overview of the algorithm CASTLE for learning MDPs modeling 
    environments with continuous stochastic dynamics.}
    \label{fig:overview}
\end{figure*}

\noindent\textbf{Overview of CASTLE.}
To increase scalability of state-of-the-art automata learning algorithms,
CASTLE makes crucial use of \emph{dimensionality reduction} and \emph{clustering}. To increase the accuracy of the learned models, CASTLE performs a combination of \emph{passive automata learning} with active sampling of new trajectories. 
Figure~\ref{fig:overview} provides an overview of CASTLE which works in two steps. 

\noindent\textbf{Step 1. - Learning an initial MDP.}
For the first step, CASTLE implements \emph{passive automata learning} 
to learn an initial MDP model of the environment. 
This step starts from a given multiset of trajectories sampled by an
existing, potentially non-optimal agent operating within 
the environment. For example, the trajectories could be collected during the
training phase of a DRL agent.
A trajectory is a sequence of observations of the environment’s state and actions chosen by the agent. This is illustrated in the 
top left of Fig.~\ref{fig:overview}.
For simple, low-dimensional, discrete applications, \emph{passive MDP learning}~\cite{DBLP:journals/ml/MaoCJNLN16} could directly be applied to learn an MDP model under the agent's decisions from the collected trajectories of the agent.
However, due to the large observation space, it is difficult for passive MDP learning to extract the relevant information and to compute a concise MDP from the original trajectories.

To overcome this issue, we \textbf{(a)} process the observations by performing \emph{dimensionality reduction} to determine relevant information in the observations and then \textbf{(b)} perform \emph{clustering} on the reduced observations.
Finally, \textbf{(c)} we use the resulting clusters as equivalence classes of the original observations and apply passive automata learning over the clusters to learn a first abstraction of the MDP under the policy of the agent. These three steps are illustrated on the bottom
of Fig.~\ref{fig:overview}.
Automata learning essentially identifies temporal 
dependencies between clusters and executed actions. It also 
splits observed clusters into different states, if they 
have different future behavior. 

\noindent\textbf{Step 2. - Fine-tuning of the model.}
In the second step, CASTLE iteratively improves the learned MDP model by
actively gathering new information
about the environment.
It is likely that the initially learned model does not accurately capture 
the environment dynamics due to employed abstraction via dimensionality reduction and 
clustering. 
To improve the model, we actively sample new trajectories to
extend our knowledge about the environment. 
For sampling the environment, we compute policies that solve the task under consideration from the learned intermediate models.
Using the newly collected traces, we gradually
increase the accuracy of learned models until the computed 
policies consistently solve the task. The right-hand side of 
Fig.~\ref{fig:overview} depicts this process.

\noindent\textbf{Main contributions.}
To the best of our knowledge, we are the first to apply automata 
learning to \emph{learn models of stochastic environments over 
continuous state space without placing assumptions on their 
unknown dynamics}. 
Having such a model makes it possible to plan ahead and to judge the agents decisions based on how likely it is that executing the agent's actions will result in an episode that successfully completes the task.
We showcase the viability of our approach by solving control tasks
from OpenAI Gym~\cite{DBLP:journals/corr/BrockmanCPSSTZ16} that serve as benchmarks for RL using policies derived from learned MDPs.

\section{Related Work}
\textsc{Alergia}~\cite{Carrasco_Oncina_1994}
was an early automata learning algorithm for 
learning probabilistic finite automata. 
\textsc{IoAlergia}\cite{DBLP:journals/corr/abs-1212-3873,DBLP:journals/ml/MaoCJNLN16} extends \textsc{Alergia} to be able to learn MDPs. 
Aichernig and Tappler subsequently applied \textsc{IoAlergia}
for probabilistic verification~\cite{DBLP:conf/rv/AichernigT17}.
These works
have in common that observations are entirely
discrete.


Instead of abstracting continuous dynamics to
discrete dynamics, Niggemann et al.~\cite{DBLP:conf/aaai/NiggemannSVMB12} and 
Medhat et al.~\cite{DBLP:conf/emsoft/MedhatRBF15}
learn hybrid automata. These automata have the
drawback that most analyses are undecidable. 
As in our approach, Kubon et al.~\cite{DBLP:conf/iberamia/KubonMR22} applied 
dimensionality reduction and passive learning to 
learn automata. However, they learn deterministic 
automata and target image classification. 

Finally, there are various approaches for learning automata 
over infinite state spaces that place restrictive assumptions 
on the environment~\cite{DBLP:conf/fm/AartsHKOV12,DBLP:conf/formats/GrinchteinJL04,DBLP:conf/icgi/VerwerWW10} .
These works have in common that the learned automata 
have (uncountable) infinite state spaces, but they can express dynamics of the environment only in a limited way.

Discretization-based approaches to learning automata 
from cyber-physical systems have been proposed 
in~\cite{DBLP:conf/pts/AichernigB0HPRR19,DBLP:conf/epew/Meinke17}, but with the general drawback that the state spaces of the learned automata explode since their automata are deterministic.
The problem of system identification~\cite{Ljung1999,DBLP:journals/tcyb/LvLHGCW18} 
addresses a similar problem,
but targets hybrid systems rather than stochastic systems
and places strong assumption on the properties of the identified models.



We perform a variant of model-based RL with the help of 
clustering. Clustering-based approaches have been proposed
for the discovery of so-called options in hierarchical model-based
RL~\cite{DBLP:journals/corr/KrishnamurthyLK16,DBLP:conf/icml/MannorMHK04,DBLP:conf/icml/MachadoBB17} that represent subtasks 
of a more complex task. 
In contrast to us, they do not 
learn environmental models. 
In a different line of research, automata learning
has been proposed to bring structure into sequences of subtasks
in hierarchical RL~~\cite{DBLP:conf/aips/0005GAMNT020,DBLP:conf/cdmake/XuWONT21,DBLP:conf/aaai/NeiderGGT0021,DBLP:conf/aips/DohmenTAB0V22,DBLP:conf/aaai/Furelos-BlancoL20,DBLP:conf/aaai/GaonB20}. These works infer discrete, deterministic automata, like reward machines, 
to enable RL with non-Markovian rewards. They serve as a high-level
specification of a complex task, rather than a representation
of the environment.

Hence, most related work from formal methods and 
AI either focuses on system verification or modeling of tasks.
Both take an agent-centric view, whereas 
we focus on modeling the environment.
Having an MDP representation of the environment 
makes it possible to analyze the agent's decision making.
To the best of our knowledge,
there is no other automata learning approach 
with the same focus and capabilities with which 
we could compare directly.

\section{Preliminaries}

\subsection{Markov Decision Processes}
Given a finite set $S$, $Dist(S)$ denotes the set of probability distributions over $S$ and $\mathit{supp}(\mu)$ for $\mu \in 
Dist(S)$ denote the support of $\mu$, i.e. the set $S' \subset S$
with $s \in S' : \mu(s) > 0$.

\noindent
A \textbf{Markov decision process (MDP)} is a tuple $\langle \mathcal{S},s_0,\mathcal{A}, \trans\rangle$ where $\mathcal{S}$ is a(n) (in)finite set of states,
 $s_0 \in Dist(\mathcal{S})$ is a distribution over initial states,
$\mathcal{A}$ is a finite set of actions,
 and
 $\trans : \mathcal{S} \times \mathcal{A} \rightarrow \Dist(\mathcal{S})$ is the probabilistic transition function.
For all $s \in \states$ the available actions are $\Act(s) = \{a \in \Act\: |\: \exists s', \trans(s, a)(s') \neq 0\}$ and we assume $|\Act(s)| \geq 1$.
The environments we consider have infinite state space.

\noindent\textbf{Trajectories.} A finite trajectory $\tau$ through an MDP is an alternating sequence 
of states and actions, i.e. 
$\tau = s_0 a_1 s_1 \cdots a_{n-1} s_{n-1} a_n s_n \in s_0 \times (\Act \times \states)^*$. 

\noindent\textbf{Policy.} 
A policy resolves the non-deterministic choice of actions in an MDP.
It is a function mapping trajectories to distributions over actions.
We consider \emph{memoryless} policies that take into account only the last
state of a trajectory, i.e., policies $\pi : \mathcal{S} \rightarrow Dist(\mathcal{A})$. A \emph{deterministic} policy $\pi$ always selects
a single action, i.e., $\pi : \mathcal{S} \rightarrow \mathcal{A}$.

\noindent\textbf{Deterministic labeled MDPs} 
are defined as MDPs
 $\mathcal{M}_L = \langle \states,s_0,\Act, \trans, \lab \rangle$
 with a finite set of states, and a unique initial state $s_0 \in 
 \mathcal{S}$, and with a labeling function $\lab : \states \rightarrow O$ mapping states to observations from a finite set $O$.
 The transition function $\trans$ must satisfy the following determinism property: 
$\forall s \in \states, \forall a \in \Act : \delta(s,a)(s') > 0 \land \delta(s,a)(s'') > 0$ implies
$s' = s''$ or $\lab(s') \neq \lab(s'')$. Given a trajectory $\tau$ in a deterministic labeled MDP $\mathcal{M}_L$,
applying the labeling function on all states of the trajectory $\tau$ 
results in a so called \emph{observation trace} $L(\tau) = L(s_0) a_1 L(s_1) \cdots a_{n-1} L(s_{n-1}) a_n L(s_n)$.
Note that due to determinism, an observation trace  $L(\tau)$ uniquely
identifies the corresponding trajectory $\tau$. 

In this paper, we use passive automata learning to compute abstract MDPs of the environment under an agent's policy in the form of deterministic labeled MDPs. The labeled MDPs
capture the information required for the agent to make its decision on how to successfully complete its task.

\subsection{Learning of MDPs}
We learn \emph{deterministic labeled MDPs} via  the algorithm \IOALERGIA~\cite{DBLP:journals/corr/abs-1212-3873,DBLP:journals/ml/MaoCJNLN16}, an adaptation of \ALERGIA{}~\cite{Carrasco_Oncina_1994}.
\IOALERGIA{}~takes a multiset $\mathcal{T}_o$ of observation traces as input.
In a first step, \IOALERGIA{}~constructs a tree that represents the observation traces by merging 
common prefixes. 
The tree has edges labeled with actions and nodes
that are labeled with observations. Each edge corresponds to 
a trace prefix with the label sequence that is visited by traversing the 
tree from the root to the node reached by
the edge.
Additionally, edges are associated
with frequencies that denote how many traces in $\mathcal{T}_o$ have 
the trace corresponding to an edge as a prefix. Normalizing these 
frequencies would already yield a tree-shaped MDP.

For generalization, the tree is transformed into an MDP with cycles through
 iterated merging of nodes. Two nodes are merged if they are compatible, i.e., 
their future behavior is sufficiently similar. For this purpose, we check
if the observations in the subtrees originating in the nodes 
are not statistically different. A parameter $\epsilon$
controls the significance level of the applied statistical tests. 
If a node is not compatible with any other  node, it is 
promoted to an MDP state. Once all pairs of nodes have been 
checked, the final deterministic labeled MDP is created by normalizing 
the frequencies on the edges to yield probability distributions for the
transition function $\trans$. In this paper, we refer to this construction
as MDP learning. 

\section{Overview of CASTLE}

\begin{figure}[t]
    \centering
    \input{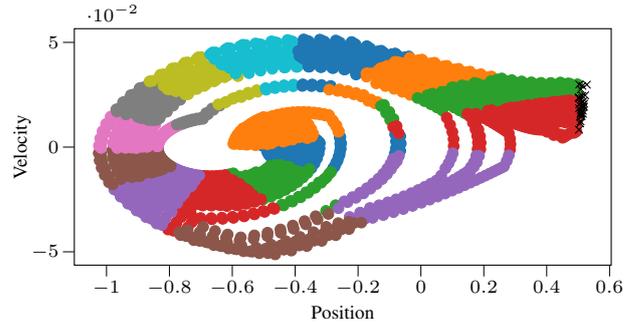} 
    \caption{States observed 
    in the Mountain Car environment. 
    }
    \label{fig:demo_example_mc}
\end{figure}


\noindent\textbf{Setting.}
We consider settings in which an agent has to perform an episodic task
in an environment modeled as an MDP $\mathcal{M} = \langle \states,s_0,\Act, \trans\rangle$. 
An episodic task is a task that has an end, i,e., ends in a terminal state
of $\mathcal{M}$.  An episode is a sequence of agent-environment interactions 
from a randomly distributed initial state of $M$ to a terminal state. 
An episode successfully completes the task to be learned if it ends in a (terminal) goal state.
If an episode ends in a bad (terminal) state, the episode fails to complete the task.

In our setting, $\mathcal{M}$ is an MDP with continuous state space $\mathcal{S} \subseteq \mathbb{R}^h$ and unknown stochastic dynamics and structure. Thus, we do not assume knowledge about the reachable states or transitions
of $\mathcal{M}$.

We are given a multiset of trajectories $\mathcal{T}$ sampled from
$\mathcal{M}$ when executing a potentially non-optimal policy on $\mathcal{M}$. 
We assume that a subset of $\mathcal{T}$ are successful trajectories and end in a goal state.  These trajectories serve as a 
starting point for passive automata learning. 

\noindent\textbf{Example.}
To illustrate the setting, consider the well-known Mountain Car 
environment available in OpenAI Gym~\cite{DBLP:journals/corr/BrockmanCPSSTZ16}, where 
the goal is to push a car up a hill. The state
space of $\mathcal{M}$ consists of two real-valued variables representing 
the x-position of the car and its velocity. 
The agent has the following three actions to interact with the environment:
accelerate to the left, accelerate to the right, and
do nothing. In the initialization of an episode, the x-position of the 
car is randomly sampled from the interval $[-0.6,-0.4]$. 
An episode is successful if the car reaches the position $0.5$ (defines the goal terminal states in $\mathcal{M}$). After $200$ time steps (defines the terminal bad states), the episode terminates as unsuccessful. 
Figure~\ref{fig:demo_example_mc} displays the 
states observed during the execution of a policy $\pi$ over $30$ episodes. 
Black x-markers indicate goal states at the x-position
$0.5$.

\noindent\textbf{Problem statement.}
Consider the setting as discussed above. The goal of our approach is to learn a concise model 
$\mathcal{M}_a = \langle \mathcal{S}_a,s_{a,0},\mathcal{A}, \trans_a, L\rangle$ in the form of an abstract deterministic labeled MDP
that models the environment $\mathcal{M}$. 
~The model $\mathcal{M}_{a}$ should be so accurate that a policy $\pi_a : \mathcal{S}_a \rightarrow \mathcal{A}$ that solves the task to be learned in $\mathcal{M}_{a}$, also successfully solves the task when being executed on $\mathcal{M}$.


\noindent\textbf{Overview of the learning process.}
Our approach consists of two phases: an initialization
and a fine-tuning phase. 
The initialization phase prepares the 
trajectories in the continuous state space so that they can be used to
learn an abstract, concise MDP model $\mathcal{M}_a$. 
Fine-tuning computes a policy based on $\mathcal{M}_a$ and uses
it to collect additional trajectories to fine-tune $\mathcal{M}_a$
with more information about the environment $\mathcal{M}$. 
The policy $\pi_a : \mathcal{S}_a \rightarrow \mathcal{A}$ for $\mathcal{M}_a$ is automatically computed 
by solving a reachability objective through
probabilistic model checking.
The computed policy $\pi_a$ selects actions that have the maximal probability of successfully completing the task.

\section{Initial Model Learning}
\label{sec:init}
The section covers the initialization phase
of CASTLE that sets up automata learning
and learns a first MDP.
For the remainder of this section, let $\mathcal{M} = \langle \mathcal{S},s_0,\mathcal{A}, \trans\rangle$, be the MDP underlying
the environment with state space $\mathcal{S} \subseteq \mathbb{R}^h$,
where $h$ is the size of the state vectors, and let $\mathcal{T}$
be a multiset of trajectories $(\mathcal{S} \times \mathcal{A})^* \times
\mathcal{S}$. 
Our goal is to learn an abstract deterministic labeled MDP
$\mathcal{M}_{a}$ serving as an initial model of $\mathcal{M}$ from the trajectories $\mathcal{T}$. 

First, our approach transforms the trajectories in $\mathcal{T}$, which are sequences of real-valued 
states and actions, into observation traces, which are sequences of 
abstract observations and actions. The transformation consists of the 
steps: (1) dimensionality reduction and scaling, (2) clustering,
and (3) additional labeling of states.
After performing these data-processing steps, we compute an initial model $\mathcal{M}_{a} = \langle \mathcal{S}_a,s_{a,0},\mathcal{A}, \trans_a, L\rangle$ using \textsc{IoAlergia} as a final fourth step. 
In the following, we discuss the individual steps in detail.

\noindent\textbf{1. Dimensionality Reduction and Scaling.}
For high-dimensional state spaces, we apply a
dimensionality reduction $sd : \mathcal{S} \rightarrow \mathcal{S}^d$ to transform $\mathcal{S}\subseteq \mathbb{R}^h$ 
to $\mathcal{S}^d \subseteq \mathbb{R}^d$ with $h>d$. We denote with $\mathcal{S}_{T}$ and 
$\mathcal{S}^d_T$ the set of all states and reduced states contained in $\mathcal{T}$.

While any standard technique like principle component analysis~\cite{MACKIEWICZ1993303} can be applied, we propose to apply a dimensionality reduction approach that works with the given trajectories in $\mathcal{T}$ instead of working only with the observed states. 
Thus, we propose to guide the selection of the 
dimensionality reduction by the actions taken in the observed states.  
For control applications, the information loss incurred by dimensionality reduction is low, if the action taken in a state can be 
predicted with
the same accuracy in $\mathbb{R}^d$ as in
$\mathbb{R}^h$.

\emph{Linear discriminant analysis} (LDA) is a natural fit for classifying states according to actions fully automatically. 
LDA can be applied to compute discriminant functions mapping states to actions such that they match the state-action pairs of the demonstration trajectories. That is, we apply LDA to learn a classifier from states to actions using the demonstrations $\mathcal{T}$. 
This enables reducing the dimensionality 
by projection to the $d$ most discriminative
axes.

Alternatively, we propose a semi-manual technique for dimensionality reduction using \emph{decision trees} (DTs). If domain knowledge is available, it can be used to extract $d$ features from the states and train DTs with a bounded size to classify
states to actions.
If the classification accuracy of DTs trained
in $\mathbb{R}^d$ (reduced states)
is similar to the accuracy in $\mathbb{R}^h$
(original states), we assume that the 
$d$ extracted features 
are sufficient for the task
at hand. The choice of the $d$ features 
allows a dimensionality reduction with a small loss of information.

Note that dimensionality reduction is 
optional and can be skipped if the number 
of dimensions is low. 
After dimensionality reduction, 
the reduced states can be further prepared for ideal clustering by applying power transformation~\cite{10.1093/biomet/87.4.954} and scaling the state data to zero mean and unit variance.

\noindent\textbf{2. Clustering.}
In the next step, CASTLE applies a clustering function $\mathit{clust} : \mathcal{S}^d \rightarrow K$ to assign cluster labels $K=\{1,\dots,k\}$ to dimensionality-reduced states.
To facilitate an application in CASTLE, the clustering approach
should enable estimating cluster membership via distances, and it should be efficient. 
CASTLE exploits the first property to simulate learned
MDPs during sampling; see Section~\ref{sec:finetune}.
One clustering approach that satisfies these requirements is the popular $k$-means algorithm~\cite{DBLP:journals/tit/Lloyd82}.


\noindent\textbf{3. Labeling.}
In the next step, CASTLE assigns a set of labels to the states in a trajectory in reduced dimensions. The labels 
are observations that are crucial for computing a policy over a learned MDP. 

We define a labeling function 
 $\mathit{lab} : \mathcal{S}^d \times \mathbb{N}_0 \rightarrow 2^O$, for a set of observations $O=\{\mathit{init}, \mathit{goal}, \mathit{bad}\}\, \cup\, K$. The function assigns 
 labels to $(s,i)$ where $s$ is a state at index
 $i$ in a trajectory.
A pair $(s,i)$ is labeled with $\mathit{init}$, if $i = 0$, i.e., $s$ is an initial state in a demonstration trajectory.
 For any other $(s,i)$, the cluster label $\mathit{clust}(s)$
 is contained in $\mathit{lab}(s,i)$.
Furthermore, 
we have $\mathit{goal}\in \mathit{lab}(s,i)$ if $s$ is a goal state or 
$\mathit{bad}\in \mathit{lab}(s,i)$ if $s$ is a bad state. Both depend on $i$, since tasks
may define a time limit for successful completion.
If additional domain knowledge is available, it can be used to assign additional labels to states, e.g., to indicate potentially dangerous situations.

\noindent\textbf{4. MDP Learning.}
To learn an MDP from trajectories $\mathcal{T}$, CASTLE transforms the trajectories into
observation traces $\mathcal{T}_O$, by sequentially applying dimensionality reduction,
scaling, clustering, and labeling. 
Given $\mathcal{T}_O$ as input,  \textsc{IoAlergia}
computes a deterministic labeled MDP $\mathcal{M}_a$, which provides
an abstract representation of the environment dynamics. 
The introduction of the 
$\mathit{init}$ label for initial states enables modeling environments where the initial states are randomly
distributed. 
Introducing the $\mathit{init}$ observation basically ignores 
the concrete initial environment 
states. Consequently, the transitions from the learned initial state (labeled $\mathit{init}$) include the distribution of initial environment states.

\noindent
\textbf{Example. }
The colors in Fig.~\ref{fig:demo_example_mc}
indicate different clusters of states derived with 
k-means and \textit{k=16} and 
the black x-markers indicate goal states. 
Thus,
the states on the right-hand side of the figure have
two labels corresponding to the clusters and 
the label $\mathit{goal}$. All other states have a single label corresponding to a cluster unless they are reached at the end of an 
episode, in which case they are labeled $\mathit{bad}$ to indicate
a timeout. Thus, states that are observed at different 
times may be labeled differently when abstracting a trajectory
to an observation trace. 
An example of an observation trace may be $\{\mathit{init}\} \cdot
\mathit{left} \cdot \{c_1\} \cdot \mathit{right}  \cdot \{c_1\} \cdots \mathit{right}
\cdot \{c_{15}, \mathit{goal}\}$. It starts with  $\mathit{init}$
and then alternates between actions and observations that include cluster labels.
The final observation includes $\mathit{goal}$ to indicate a successful episode.
With this information, \textsc{IoAlergia}
is able to learn temporal dependencies between 
observations and actions.

\section{Model Fine-Tuning}
\label{sec:finetune}
The fine-tuning phase of CASTLE incrementally improves the learned labeled MDP $\mathcal{M}_a$ that models the environment $\mathcal{M}$.
In a nutshell, the fine-tuning phase iteratively (1) computes a policy that is able to solve the task in $\mathcal{M}_a$ via probabilistic model checking,
(2) uses the policy to sample new trajectories, and (3) learns a new, improved
model with the extended multiset of trajectories.

The fine-tuning phase is based on the approach proposed by Aichernig and Tappler~\cite{DBLP:conf/rv/AichernigT17}. 
In contrast to the original approach, which works
solely on abstract observations, our fine-tuning approach takes the concrete state space and the uncertainties stemming from clustering
into account. 
In the following, we discuss the individual steps of our fine-tuning approach in detail.

\noindent
\textbf{1. Policy Computation.}
Given $\mathcal{M}_a = \langle \mathcal{S}_a,s_{a,0},\mathcal{A}, \trans_a, L\rangle$, our goal is to compute 
a (deterministic) policy $\pi_a : \mathcal{S}_a \rightarrow \mathcal{A}$
that maximizes the probability to complete the task successfully, i.e., reaching a goal state in $\mathcal{M}_a$.
Thus, for any state $s_a\in \mathcal{S}_a$, the policy $\pi_a$ picks the action $a_{\pi}=\pi_a(s_a)$ that maximizes the probability of reaching a goal state. Such policies can be automatically computed via probabilistic model checking (using tools like \textsc{Prism}~\cite{prism}). 
Let $p_{s,a} = P_\mathrm{max}(\mathbf{F}\ \mathit{goal}, s_a, a)$ be the maximal probability of reaching a goal state from state $s_a$ when executing action $a$, where $\mathbf{F}$ denotes the \emph{eventually} operator.
For any state $s_a\in \mathcal{S}_a$, we have $a_{\pi} = \pi(s_a)$ with $a_{\pi}=\max_{a\in \mathcal{A}} p_{s_a,a}$. 





\noindent
\textbf{2. Sampling.}
In the next step of the iterative model refinement phase, CASTLE uses the policy $\pi_a: \mathcal{S}_a \rightarrow \mathcal{A}$ over the current $\mathcal{M}_a$ to sample additional trajectories in $\mathcal{M}$.
The newly sampled trajectories are added 
to the existing trajectories $\mathcal{T}$ and are used in the next step to improve the accuracy of $\mathcal{M}_a$. 


\noindent\textbf{Overview of sampling of trajectories.}
For the sampling, the learned MDP 
$\mathcal{M}_a = \langle \mathcal{S}_a,s_{a,0},\mathcal{A}, \trans_a, L\rangle$ is simulated in parallel with the environment $\mathcal{M} = \langle \mathcal{S},s_{0},\mathcal{A}, \trans\rangle$. The actions are selected via the policy $\pi_a: \mathcal{S}_a \rightarrow \mathcal{A}$, computed in the previous step.

During sampling, the MDP $\mathcal{M}_a$ is treated similarly to a partially observable MDP (POMDP).
To account for inaccuracies in $\mathcal{M}_a$, 
we adopt the notion of belief states (belief for short).
A belief $B$ is a distribution over the states $\mathcal{S}_a$, i.e., 
at any time step of the current episode, the $\mathcal{M}_a$ is in state $s_a \in \mathcal{S}_a$ with probability $B(s_a)$.
The belief update is both based on the 
structure of learned MDP $\mathcal{M}_a$ as well as the 
environment state reached after a step. 
In time step $i$, after having taken an action $act$ and the environment having moved to a state $s'\in \mathcal{S}$ in cluster $k'$, 
we update the belief $B$ to $B'$
to include states $s_a'$ with $\trans_a(s_a,act,s_a') > 0$ for $s_a \in \mathit{supp}{B}$. That is, we move to states
reachable in $\mathcal{M}_a$.
The probabilities $B'(s_a')$ are the product 
of $B(s_a)$, i.e., the previous state probability, and 
a term that is inversely
proportional to the distance between the cluster
centroid of $k'$ (the reached cluster)
and the centroid of the cluster in $L(s_a')$.

Consider the following scenario to see why keeping 
track of a single state $s_a\in \mathcal{S}_a$ is not sufficient.
Suppose that in the current episode, the environment $\mathcal{M}$ is in a state $s\in \mathcal{S}$ and the learned model $\mathcal{M}_a$ is in a state $s_a\in \mathcal{S}_a$.
In the next time step, the environment moves from state
$s$ to $s'$ via action $act$. The corresponding cluster labels of the states $s$ and $s'$ are $k$ and $k'$, respectively. 
Ideally, in the learned model $\mathcal{M}_a$,
there is a unique $s'_a$ corresponding to $s'$
identified by $\trans_a(s_a,a,s_a') > 0$ 
with $k' \in L(s_a')$.
However, since the learned models are not perfectly accurate,
especially during early iterations, this is not 
always the case.

\begin{algorithm}[t]
\caption{Sampling with a policy computed from a learned MDP.}
\label{alg:sample}
\begin{algorithmic}[1]
\Require{$\mathcal{M} = \langle \mathcal{S},s_{0},\mathcal{A}, \trans\rangle$, model $\mathcal{M}_a = \langle \mathcal{S}_a,s_{a,0},\mathcal{A}, \trans_a, L\rangle$, policy $\pi : \mathcal{S}_a \rightarrow \mathcal{A}$, clusters $K$, max. belief size $b_n$}
\Ensure{A sampled trajectory $\tau$}
\State $s \in \mathcal{S} \gets \texttt{reset()}$ \label{line:reset}
\State $\tau \gets s$
\State $B \gets \{s_{a,0} \mapsto 1\}$ \label{line:belief_init}
\While{$s$ is not terminal}
\label{line:while}
\State $ba \gets \{a \mapsto \sum_{s_a \in \mathcal{S}_a,\pi_a(s_a) = act} B(s_a) 
|act \in \mathcal{A} \}$ \label{line:action_distr}
\State sample $act \sim ba$ 
\State $s \gets \texttt{step}(act)$
\State $\tau \gets \tau \cdot act \cdot s$
\State $dists \gets \{k \mapsto dist(sd(s),
centroid(k)) | k \in K \}$
\label{line:cdf}
\State $cDistr \gets \textit{cdf. of } \mathcal{N}(mean(dists); 
stddev(dists)^2)$
\State $B' \gets \{s_a \mapsto 0|s_a \in \mathcal{S}_a\}$
\For{$s_a \in \mathit{supp}(B)$, $s_a' \in \mathit{supp}(\trans_a(s_a,act))$} \label{line:forloop_reachable}
\State $k' \gets L(s_a') \cap K$ 
\State $B'(s_a') \gets B'(s_a') +  B(s_a) \cdot (1-cDistr(k'))$ \label{line:individual_belief_update}
\EndFor
\State $B' \gets \{s_a \mapsto p \in B'|p \textit{ is in the }b_n\textit{ largest values of } B'(\cdot)\}$
\label{line:discard}
\State $B' \gets \{s_a \mapsto \frac{p}{\sum_{s_a'}B(s_a')}|s_a \mapsto p  \in B'\}$
\label{line:normalize}
\EndWhile
\State \Return $\tau$
\end{algorithmic}
\end{algorithm}

\noindent\textbf{Algorithm for sampling of trajectories.}
Algorithm~\ref{alg:sample} formalizes our approach to sample
a trajectory from $\mathcal{M}$ using a policy computed from a learned model $\mathcal{M}_a$.
We  follow the OpenAI Gym~\cite{DBLP:journals/corr/BrockmanCPSSTZ16} conventions and use the operations $\texttt{reset}$ and $\texttt{step}$ to change the current state $s$ of the environment $\mathcal{M}$. The function
$\texttt{reset}$ resets $\mathcal{M}$ to an
initial state and returns this state. The function 
$\texttt{step}$ takes an action $a$ as input, executes $a$, and
returns the reached state $s'$.

In Algo.~\ref{alg:sample}, the Lines \ref{line:reset} to \ref{line:belief_init}
perform initialization steps by resetting 
the environment, adding the initial state to the sampled 
trajectory $\tau$, and initializing the belief $B$ 
to include only the initial state of 
$\mathcal{M}_a$. 

We then sample experiences from $\mathcal{M}$
until reaching a terminal state (Line \ref{line:while}).
Line \ref{line:action_distr} transforms
the belief state into a distribution over actions
by mapping states to actions chosen by the policy $\pi_a$. 
The next $3$ lines sample an action, 
execute it in the environment, and append the
new state-action pair to the trajectory $\tau$.
Hence, the combination of the current
belief and a deterministic policy $\pi_a$ yields
a probabilistic policy.

Starting in Line \ref{line:cdf}, we begin 
the update of the belief state. 
First, we compute the Euclidean distances from all cluster 
centroids to the current environment state in 
reduced dimensions. We then 
fit a normal distribution over distances, which 
we empirically found to be a good fit.
After that, we initialize the next belief 
state $B'$ and iterate over all states
reachable in $\mathcal{M}_a$ (Line~\ref{line:forloop_reachable}).
In Line~\ref{line:individual_belief_update}, 
we add the contribution of $s_a'$ to the next belief $B'$
as the product of the previous belief $B(s_a)$
and the inverse of the distance probability of 
the cluster $k'$ labeling $s_a'$. We use 
the inverse to favor short distances and 
to ignore $\trans(s_a,a)$ so that we only consider the
actual environment information for the belief update.
In Line \ref{line:discard} we discard states with low
probability. 
Finally, Line
\ref{line:normalize} normalizes the belief to a distribution.

\noindent
\textbf{3. Learning and Stopping.}
In each iteration of the fine-tuning phase, we sample $n_\mathrm{samples}$ additional trajectories from $\mathcal{M}$, as outlined above.
To learn a more accurate model $\mathcal{M}_a$ from the additional information, CASTLE takes the newly collected trajectories 
$\mathcal{T}^{new}$ and transforms them into observation traces $\mathcal{T}^{new}_O$. This is done by sequentially applying dimensionality reduction, scaling, clustering, and labeling
as outlined in Sect.~\ref{sec:init}.
CASTLE adds the new observation traces to the existing multiset of traces $\mathcal{T}_O$ and learns a labeled MDP $\mathcal{M}_a$ with \textsc{IoAlergia}.
After computing a new learned model $\mathcal{M}_a$, CASTLE returns 
to the policy computation step. 

As stopping criteria for the iteration, one can stop either after a fixed number of iterations or 
upon reaching a goal state a specified number 
of times in the current iteration.


\section{Experiments}
\label{sec:experiments}

\begin{table*}[t]
\resizebox{\textwidth}{!}{
\begin{tabular}{|l|l|l|l|l|l|l|l|l|l|}
\hline
    Environment &
  \begin{tabular}[c]{@{}l@{}}Demonstration \\ Trajectories\end{tabular} &
  \begin{tabular}[c]{@{}l@{}}Demonstration \\ Timesteps\end{tabular} &
  \begin{tabular}[c]{@{}l@{}}Dimensionality\\ Reduction\end{tabular} &
  Clusters &
  \begin{tabular}[c]{@{}l@{}}Fine-tuning \\ Iterations\end{tabular} &
  \begin{tabular}[c]{@{}l@{}}Episodes Per\\ Fine-tuning\end{tabular} &
  \begin{tabular}[c]{@{}l@{}}Final Model\\ Size\end{tabular} &
  \begin{tabular}[c]{@{}l@{}}Best Policy \\ Reward\end{tabular} &
  \begin{tabular}[c]{@{}l@{}}Goal \\ Reached\end{tabular} \\ \hline

Acrobot &
  2500 &
  2.1 $\times 10^5$ &
  \begin{tabular}[c]{@{}l@{}}Manual Mapper\\ LDA + Power Transformer\end{tabular} &
  256 &
  25 &
  50 &
  \begin{tabular}[c]{@{}l@{}}472\\ 482\end{tabular} &
  \begin{tabular}[c]{@{}l@{}}-113 $\pm$ 41\\ -156 $\pm$ 63\end{tabular} &
  \begin{tabular}[c]{@{}l@{}}42\% | 82\%\\ 4\% | 35\%\end{tabular} \\ \hline
Lunar Lander &
  2500 &
  5.6  $\times 10^5$ &
  \begin{tabular}[c]{@{}l@{}}Manual Mapper\\ LDA + Power Transformer\end{tabular} &
  512 &
  25 &
  50 &
  \begin{tabular}[c]{@{}l@{}}1122\\ 1100\end{tabular} &
  \begin{tabular}[c]{@{}l@{}}213 $\pm$ 80\\ 103 $\pm$ 138\end{tabular} &
  \begin{tabular}[c]{@{}l@{}}81\%\\ 35\%\end{tabular} \\ \hline
Mountain Car &
  2500 &
  3 $\times 10^5$ &
  Power Transformer &
  256 &
  25 &
  50 &
  363 &
  -136 $\pm$ 28 &
  42\% \\ \hline
Cartpole &
  2500 &
  5 $\times 10^5$ &
  Power Transformer &
  128 &
  15 &
  50 &
  213 &
  195 $\pm$ 18 &
  88\% \\ \hline
\end{tabular}%
}
\caption{Parameterized results for all environments. All values are averages from five experiment runs.}
\label{tab:experiment_data}
\end{table*}
We applied CASTLE to four well-known RL applications~\cite{DBLP:journals/corr/BrockmanCPSSTZ16}: (1) Acrobot, (2) Lunar Lander, (3) Mountain Car, and (4) Cartpole. All reported results are average values over five experiment runs per environment. Pretrained \textit{Stable-Baselines3}~\cite{stable-baselines3} RL agents that achieved high reward on their respective tasks were downloaded from HuggingFace\footnote{\url{https://huggingface.co/}}. Model learning was done with~\textsc{AALpy}~\cite{DBLP:journals/isse/MuskardinAPPT22}. For the computation of the policies over the learned models we used the model checker \textsc{Prism}~\cite{prism}.
All experiments were conducted on a laptop with an Intel\textsuperscript{\textregistered} Core\texttrademark i7-11800H CPU at 2.3 GHz with 32 GB of RAM. Source code and detailed instructions to reproduce our experiments are available in the supplementary material.

\noindent\textbf{Results.} Table~\ref{tab:experiment_data} contains the parameterized results for each experiment. The ``Demonstration Trajectories'' column gives the total number of episodes/trajectories, which we sampled 
using a DRL agent downloaded from the \url{https://huggingface.co/}
library. The column ``Demonstration Timesteps'' lists the timesteps performed on the environment during the sampling of demonstrations, i.e., the combined length of the demonstration trajectories. 

Please note the compact size of the learned models 
in the column ``Final Model Size''.
In all experiments, each environment state was observed only once (due to the continuous state space). The total number of observed states is thus equal to the number of timesteps. Therefore, if we would directly apply automata learning, the learned MDP would have [2.1-5.6]$\times 10^5$ states.
The learned MDPs of CASTLE have [213-1122] states.
Next, we discuss the results of the individual experiments in more detail.

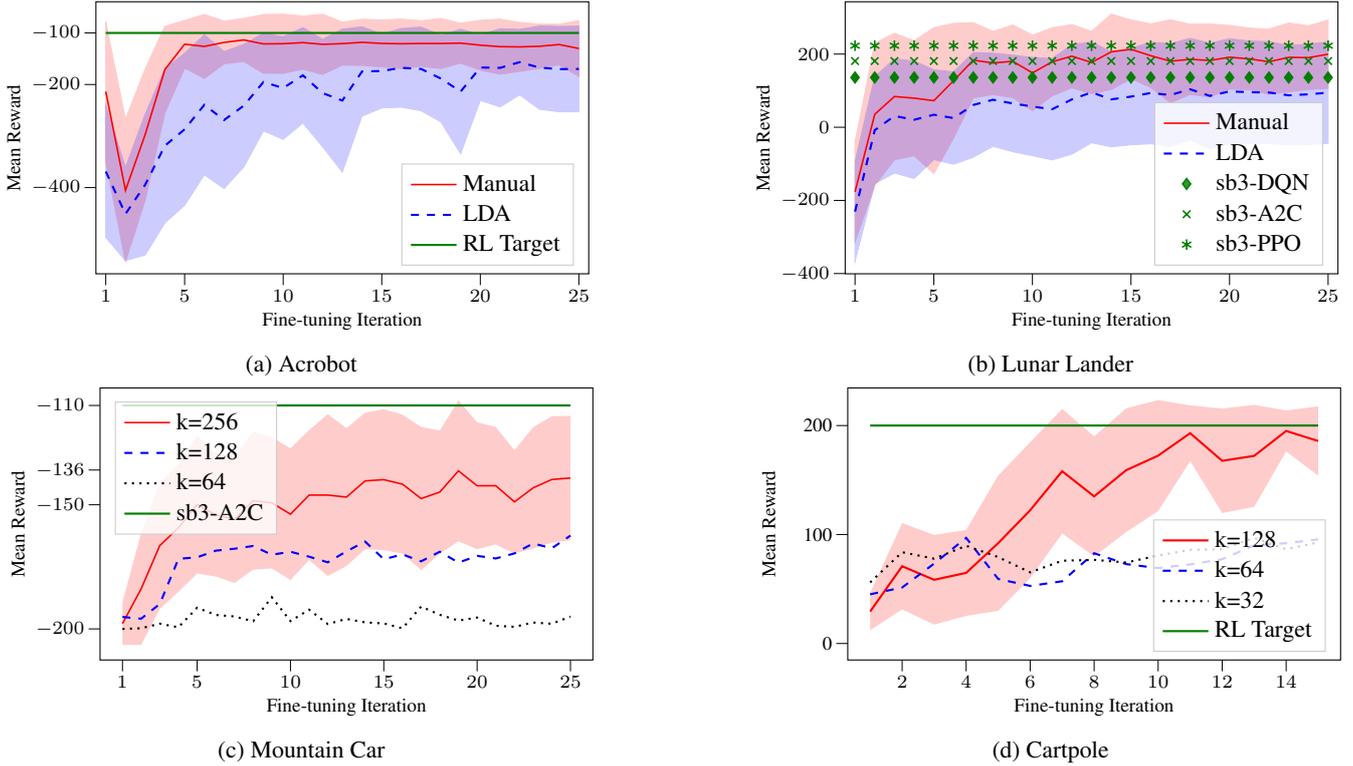
\begin{figure*}[t]
  \begin{subfigure}{0.45\textwidth}
    \centering
\begin{tikzpicture}[]

\definecolor{darkgray176}{RGB}{176,176,176}
\definecolor{green01270}{RGB}{0,127,0}
\definecolor{lightgray204}{RGB}{204,204,204}

\begin{axis}[
legend cell align={left},
legend style={
  fill opacity=0.8,
  draw opacity=1,
  text opacity=1,
  at={(0.97,0.03)},
  anchor=south east,
  draw=lightgray204
},
tick align=outside,
tick pos=left,
x grid style={darkgray176},
xlabel={Fine-tuning Iteration},
xmin=0.5, xmax=25.5,
xtick style={color=black},
xtick={1,5,10,15, 20, 25},
y grid style={darkgray176},
ylabel={Mean Reward},
ytick={-400, -200, -100},
ymin=-567.219931754423, ymax=-39.4787530326523,
ytick style={color=black},
xlabel style={yshift=0.5ex},
ylabel style={yshift=-0.5ex},
height=.22\textheight,
width=\columnwidth,
]
\path [draw=red, fill=red, opacity=0.2]
(axis cs:1,-78.5158961224853)
--(axis cs:1,-348.776103877515)
--(axis cs:2,-543.231696357979)
--(axis cs:3,-424.128100763137)
--(axis cs:4,-254.688778187906)
--(axis cs:5,-167.926872279908)
--(axis cs:6,-188.166161905092)
--(axis cs:7,-159.90176963447)
--(axis cs:8,-155.253078788634)
--(axis cs:9,-178.224484880618)
--(axis cs:10,-178.305011570904)
--(axis cs:11,-170.337408932822)
--(axis cs:12,-180.482062759752)
--(axis cs:13,-162.437410054902)
--(axis cs:14,-162.938037873013)
--(axis cs:15,-165.898455855722)
--(axis cs:16,-168.944793727445)
--(axis cs:17,-164.545626987397)
--(axis cs:18,-169.100788073375)
--(axis cs:19,-158.480401115185)
--(axis cs:20,-171.973240666827)
--(axis cs:21,-179.937687564375)
--(axis cs:22,-174.543152324379)
--(axis cs:23,-168.853246151369)
--(axis cs:24,-161.937851015028)
--(axis cs:25,-184.374597767839)
--(axis cs:25,-75.977402232161)
--(axis cs:25,-75.977402232161)
--(axis cs:24,-82.9541489849718)
--(axis cs:23,-82.8347538486314)
--(axis cs:22,-79.5208476756206)
--(axis cs:21,-73.2063124356255)
--(axis cs:20,-75.4507593331729)
--(axis cs:19,-81.0515988848153)
--(axis cs:18,-71.7512119266249)
--(axis cs:17,-76.2463730126035)
--(axis cs:16,-73.0272062725554)
--(axis cs:15,-74.6775441442784)
--(axis cs:14,-73.4339621269872)
--(axis cs:13,-78.9745899450984)
--(axis cs:12,-63.6099372402476)
--(axis cs:11,-67.1745910671782)
--(axis cs:10,-63.4669884290964)
--(axis cs:9,-64.2195151193816)
--(axis cs:8,-71.8629212113662)
--(axis cs:7,-76.9302303655298)
--(axis cs:6,-64.1058380949079)
--(axis cs:5,-75.7371277200923)
--(axis cs:4,-87.0632218120939)
--(axis cs:3,-169.791899236863)
--(axis cs:2,-267.564303642022)
--(axis cs:1,-78.5158961224853)
--cycle;

\path [draw=blue, fill=blue, opacity=0.2]
(axis cs:1,-240.916498247017)
--(axis cs:1,-496.571501752983)
--(axis cs:2,-541.531579123587)
--(axis cs:3,-530.885234668133)
--(axis cs:4,-468.503820947631)
--(axis cs:5,-434.338503851371)
--(axis cs:6,-375.198922795504)
--(axis cs:7,-401.799591962793)
--(axis cs:8,-360.031850005793)
--(axis cs:9,-290.056764534957)
--(axis cs:10,-306.450186599742)
--(axis cs:11,-274.773636725309)
--(axis cs:12,-320.866463796776)
--(axis cs:13,-369.5926677166)
--(axis cs:14,-261.385870202354)
--(axis cs:15,-245.587231018885)
--(axis cs:16,-243.933467174337)
--(axis cs:17,-251.071041886126)
--(axis cs:18,-272.569684124042)
--(axis cs:19,-334.4713438071)
--(axis cs:20,-230.623397207614)
--(axis cs:21,-243.678989653302)
--(axis cs:22,-220.111676613816)
--(axis cs:23,-248.13474704486)
--(axis cs:24,-252.931538969558)
--(axis cs:25,-252.995966835027)
--(axis cs:25,-86.6880331649728)
--(axis cs:25,-86.6880331649728)
--(axis cs:24,-87.4044610304421)
--(axis cs:23,-86.4052529551396)
--(axis cs:22,-93.1243233861837)
--(axis cs:21,-91.7730103466983)
--(axis cs:20,-103.292602792386)
--(axis cs:19,-92.8526561928995)
--(axis cs:18,-104.402315875958)
--(axis cs:17,-88.7009581138736)
--(axis cs:16,-91.2745328256627)
--(axis cs:15,-102.048768981115)
--(axis cs:14,-88.1741297976458)
--(axis cs:13,-92.7073322833998)
--(axis cs:12,-112.493536203224)
--(axis cs:11,-89.618363274691)
--(axis cs:10,-107.293813400258)
--(axis cs:9,-100.379235465043)
--(axis cs:8,-122.184149994207)
--(axis cs:7,-136.140408037207)
--(axis cs:6,-103.553077204496)
--(axis cs:5,-139.621496148629)
--(axis cs:4,-168.860179052369)
--(axis cs:3,-257.766765331867)
--(axis cs:2,-360.784420876413)
--(axis cs:1,-240.916498247017)
--cycle;

\addplot [semithick, red]
table {%
1 -213.646
2 -405.398
3 -296.96
4 -170.876
5 -121.832
6 -126.136
7 -118.416
8 -113.558
9 -121.222
10 -120.886
11 -118.756
12 -122.046
13 -120.706
14 -118.186
15 -120.288
16 -120.986
17 -120.396
18 -120.426
19 -119.766
20 -123.712
21 -126.572
22 -127.032
23 -125.844
24 -122.446
25 -130.176
};
\addlegendentry{Manual}
\addplot [dashed, thick, blue]
table {%
1 -368.744
2 -451.158
3 -394.326
4 -318.682
5 -286.98
6 -239.376
7 -268.97
8 -241.108
9 -195.218
10 -206.872
11 -182.196
12 -216.68
13 -231.15
14 -174.78
15 -173.818
16 -167.604
17 -169.886
18 -188.486
19 -213.662
20 -166.958
21 -167.726
22 -156.618
23 -167.27
24 -170.168
25 -169.842
};
\addlegendentry{LDA}
\addplot [thick, green01270]
table {%
1 -100
2 -100
3 -100
4 -100
5 -100
6 -100
7 -100
8 -100
9 -100
10 -100
11 -100
12 -100
13 -100
14 -100
15 -100
16 -100
17 -100
18 -100
19 -100
20 -100
21 -100
22 -100
23 -100
24 -100
25 -100
};
\addlegendentry{RL Target}
\end{axis}

\end{tikzpicture}
    \caption{Acrobot}
    \label{fig:acrobot}
  \end{subfigure}
  \hfill
  \begin{subfigure}{0.45\textwidth}
    \centering
\begin{tikzpicture}

\definecolor{darkgray176}{RGB}{176,176,176}
\definecolor{green01270}{RGB}{0,127,0}
\definecolor{lightgray204}{RGB}{204,204,204}

\begin{axis}[
legend cell align={left},
legend style={
  fill opacity=0.8,
  draw opacity=1,
  text opacity=1,
  at={(0.97,0.03)},
  anchor=south east,
  draw=lightgray204
},
tick align=outside,
tick pos=left,
x grid style={darkgray176},
xlabel={Fine-tuning Iteration},
xmin=0.5, xmax=25.5,
xtick style={color=black},
xtick={1,5,10,15, 20, 25},
y grid style={darkgray176},
ylabel={Mean Reward},
ymin=-400.631927254507, ymax=343.106383566032,
ytick style={color=black},
xlabel style={yshift=0.5ex},
ylabel style={yshift=-0.5ex},
height=.22\textheight,
width=\columnwidth,
]
\path [draw=red, fill=red, opacity=0.2]
(axis cs:1,-40.6688419813521)
--(axis cs:1,-312.389044548329)
--(axis cs:2,-159.196779909548)
--(axis cs:3,-89.0551004832195)
--(axis cs:4,-77.5471744296415)
--(axis cs:5,-126.937568878855)
--(axis cs:6,-29.391952397895)
--(axis cs:7,80.6346559221675)
--(axis cs:8,89.1940021672316)
--(axis cs:9,81.2166432444124)
--(axis cs:10,45.1871540201248)
--(axis cs:11,85.8507111865844)
--(axis cs:12,107.985825506617)
--(axis cs:13,92.0477118659367)
--(axis cs:14,103.151066892692)
--(axis cs:15,132.652442236227)
--(axis cs:16,108.869929623278)
--(axis cs:17,67.296254950874)
--(axis cs:18,89.8701871546032)
--(axis cs:19,82.8664762581913)
--(axis cs:20,104.474294633538)
--(axis cs:21,102.643775477636)
--(axis cs:22,89.8948329802371)
--(axis cs:23,97.0096797948915)
--(axis cs:24,103.151177296213)
--(axis cs:25,106.309661692656)
--(axis cs:25,293.00222918316)
--(axis cs:25,293.00222918316)
--(axis cs:24,277.479570573705)
--(axis cs:23,285.164938118398)
--(axis cs:22,268.359448711617)
--(axis cs:21,271.252794835945)
--(axis cs:20,278.410894520156)
--(axis cs:19,282.338137503358)
--(axis cs:18,281.590684044713)
--(axis cs:17,293.967308271457)
--(axis cs:16,282.583008795827)
--(axis cs:15,293.523639755898)
--(axis cs:14,309.300096710553)
--(axis cs:13,262.098636920548)
--(axis cs:12,280.681996768222)
--(axis cs:11,271.975623354286)
--(axis cs:10,251.966629016884)
--(axis cs:9,279.480159872637)
--(axis cs:8,262.296355931651)
--(axis cs:7,286.164938054397)
--(axis cs:6,284.050245018985)
--(axis cs:5,272.193162106738)
--(axis cs:4,236.64232316501)
--(axis cs:3,257.028271614392)
--(axis cs:2,229.783530240271)
--(axis cs:1,-40.6688419813521)
--cycle;

\path [draw=blue, fill=blue, opacity=0.2]
(axis cs:1,-94.4556471128949)
--(axis cs:1,-366.825640399028)
--(axis cs:2,-154.341600098222)
--(axis cs:3,-125.21873450214)
--(axis cs:4,-140.027709757982)
--(axis cs:5,-88.2949233191734)
--(axis cs:6,-101.237424074922)
--(axis cs:7,-81.874543127662)
--(axis cs:8,-52.2629576779233)
--(axis cs:9,-66.8905943435589)
--(axis cs:10,-77.7693239227643)
--(axis cs:11,-89.2274731152529)
--(axis cs:12,-73.4743251537915)
--(axis cs:13,-42.3577653189128)
--(axis cs:14,-53.3749593048031)
--(axis cs:15,-47.9494563705559)
--(axis cs:16,-45.0785243235094)
--(axis cs:17,-52.5936916217611)
--(axis cs:18,-35.0755787045689)
--(axis cs:19,-58.834189706912)
--(axis cs:20,-45.6560645176667)
--(axis cs:21,-44.152854568451)
--(axis cs:22,-44.138312263289)
--(axis cs:23,-48.5841473309963)
--(axis cs:24,-47.241572598915)
--(axis cs:25,-43.7177587160135)
--(axis cs:25,232.909677757308)
--(axis cs:25,232.909677757308)
--(axis cs:24,227.397997788561)
--(axis cs:23,223.101064979929)
--(axis cs:22,234.249279329544)
--(axis cs:21,235.684868918557)
--(axis cs:20,241.820540110795)
--(axis cs:19,230.189916406468)
--(axis cs:18,242.811072222512)
--(axis cs:17,228.772718071324)
--(axis cs:16,232.429582347469)
--(axis cs:15,215.559060347363)
--(axis cs:14,205.700365561867)
--(axis cs:13,233.208452682304)
--(axis cs:12,225.613561006519)
--(axis cs:11,187.703414155689)
--(axis cs:10,190.905766465367)
--(axis cs:9,197.156434858576)
--(axis cs:8,202.62613172147)
--(axis cs:7,203.85626147259)
--(axis cs:6,151.428520393475)
--(axis cs:5,156.930304572471)
--(axis cs:4,181.031787805561)
--(axis cs:3,186.796486806252)
--(axis cs:2,138.289647478824)
--(axis cs:1,-94.4556471128949)
--cycle;

\addplot [semithick, red]
table {%
1 -176.528943264841
2 35.2933751653614
3 83.9865855655864
4 79.5475743676844
5 72.627796613942
6 127.329146310545
7 183.399796988282
8 175.745179049441
9 180.348401558525
10 148.576891518504
11 178.913167270435
12 194.33391113742
13 177.073174393242
14 206.225581801623
15 213.088040996063
16 195.726469209552
17 180.631781611166
18 185.730435599658
19 182.602306880775
20 191.442594576847
21 186.94828515679
22 179.127140845927
23 191.087308956645
24 190.315373934959
25 199.655945437908
};
\addlegendentry{Manual}
\addplot [dashed, thick, blue]
table {%
1 -230.640643755962
2 -8.02597630969901
3 30.7888761520561
4 20.5020390237892
5 34.3176906266487
6 25.0955481592762
7 60.9908591724642
8 75.1815870217734
9 65.1329202575085
10 56.5682212713014
11 49.2379705202178
12 76.0696179263636
13 95.4253436816957
14 76.1627031285318
15 83.8048019884035
16 93.6755290119797
17 88.0895132247816
18 103.867746758972
19 85.6778633497782
20 98.082237796564
21 95.7660071750527
22 95.0554835331274
23 87.2584588244662
24 90.0782125948232
25 94.5959595206475
};
\addlegendentry{LDA}
\addplot [semithick, green01270, mark=diamond*, mark size=2, mark options={solid}, only marks]
table {%
1 136
2 136
3 136
4 136
5 136
6 136
7 136
8 136
9 136
10 136
11 136
12 136
13 136
14 136
15 136
16 136
17 136
18 136
19 136
20 136
21 136
22 136
23 136
24 136
25 136
};
\addlegendentry{sb3-DQN}
\addplot [semithick, green01270, mark=x, mark size=2, mark options={solid}, only marks]
table {%
1 181
2 181
3 181
4 181
5 181
6 181
7 181
8 181
9 181
10 181
11 181
12 181
13 181
14 181
15 181
16 181
17 181
18 181
19 181
20 181
21 181
22 181
23 181
24 181
25 181
};
\addlegendentry{sb3-A2C}
\addplot [semithick, green01270, mark=asterisk, mark size=2, mark options={solid}, only marks]
table {%
1 223
2 223
3 223
4 223
5 223
6 223
7 223
8 223
9 223
10 223
11 223
12 223
13 223
14 223
15 223
16 223
17 223
18 223
19 223
20 223
21 223
22 223
23 223
24 223
25 223
};
\addlegendentry{sb3-PPO}
\end{axis}

\end{tikzpicture}
    \caption{Lunar Lander}
    \label{fig:lunar_lander}
  \end{subfigure}
  \hfill
  \begin{subfigure}{0.45\textwidth}
    \centering
\begin{tikzpicture}

\definecolor{darkgray176}{RGB}{176,176,176}
\definecolor{darkorange25512714}{RGB}{255,127,14}
\definecolor{forestgreen4416044}{RGB}{44,160,44}
\definecolor{green01270}{RGB}{0,127,0}
\definecolor{lightgray204}{RGB}{204,204,204}
\definecolor{steelblue31119180}{RGB}{31,119,180}

\begin{axis}[
legend cell align={left},
legend style={
  fill opacity=0.8,
  draw opacity=1,
  text opacity=1,
  at={(0.03,0.95)},
  anchor=north west,
  draw=lightgray204,
  font=\small
},
tick align=outside,
tick pos=left,
x grid style={darkgray176},
xlabel={Fine-tuning Iteration},
xmin=-0.2, xmax=26.2,
xtick style={color=black},
xtick={1,5,10,15, 20, 25},
y grid style={darkgray176},
ylabel={Mean Reward},
ymin=-212.515219335616, ymax=-102.952772469181,
ytick style={color=black},
ytick={-200, -150,  -136, -110},
xlabel style={yshift=0.5ex},
ylabel style={yshift=-0.5ex},
height=.22\textheight,
width=\columnwidth,
]
\path [fill=red, fill opacity=0.2]
(axis cs:1,-189.268948614605)
--(axis cs:1,-206.435051385395)
--(axis cs:2,-206.479988064883)
--(axis cs:3,-191.956643567602)
--(axis cs:4,-185.095440406524)
--(axis cs:5,-177.727340572676)
--(axis cs:6,-178.897119581822)
--(axis cs:7,-181.517691362459)
--(axis cs:8,-176.138828551824)
--(axis cs:9,-175.663370805868)
--(axis cs:10,-180.329152709492)
--(axis cs:11,-172.293939947244)
--(axis cs:12,-178.728534582116)
--(axis cs:13,-174.67765700356)
--(axis cs:14,-167.879901984017)
--(axis cs:15,-168.269586307846)
--(axis cs:16,-169.691043738054)
--(axis cs:17,-176.343331135707)
--(axis cs:18,-169.693391901767)
--(axis cs:19,-164.747116309617)
--(axis cs:20,-167.914255684376)
--(axis cs:21,-166.005137567145)
--(axis cs:22,-169.642652087358)
--(axis cs:23,-167.662294294461)
--(axis cs:24,-165.305732934898)
--(axis cs:25,-164.26561906272)
--(axis cs:25,-114.15038093728)
--(axis cs:25,-114.15038093728)
--(axis cs:24,-114.350267065102)
--(axis cs:23,-118.705705705539)
--(axis cs:22,-128.029347912643)
--(axis cs:21,-118.650862432855)
--(axis cs:20,-116.781744315624)
--(axis cs:19,-107.932883690383)
--(axis cs:18,-120.066608098233)
--(axis cs:17,-118.656668864293)
--(axis cs:16,-113.700956261946)
--(axis cs:15,-111.498413692154)
--(axis cs:14,-112.896098015983)
--(axis cs:13,-119.11434299644)
--(axis cs:12,-113.463465417884)
--(axis cs:11,-119.906060052756)
--(axis cs:10,-127.246847290508)
--(axis cs:9,-122.808629194132)
--(axis cs:8,-120.517171448176)
--(axis cs:7,-130.274308637541)
--(axis cs:6,-129.350880418178)
--(axis cs:5,-122.064659427324)
--(axis cs:4,-133.528559593476)
--(axis cs:3,-140.891356432398)
--(axis cs:2,-161.304011935117)
--(axis cs:1,-189.268948614605)
--cycle;

\addplot [semithick, red]
table {%
1 -197.852
2 -183.892
3 -166.424
4 -159.312
5 -149.896
6 -154.124
7 -155.896
8 -148.328
9 -149.236
10 -153.788
11 -146.1
12 -146.096
13 -146.896
14 -140.388
15 -139.884
16 -141.696
17 -147.5
18 -144.88
19 -136.34
20 -142.348
21 -142.328
22 -148.836
23 -143.184
24 -139.828
25 -139.208
};
\addlegendentry{k=256}
\addplot [dashed, thick, blue]
table {%
1 -195.188
2 -195.896
3 -189.964
4 -171.764
5 -171.16
6 -168.48
7 -167.708
8 -166.604
9 -170.132
10 -168.912
11 -170.904
12 -173.184
13 -168.76
14 -164.82
15 -171.956
16 -169.98
17 -172.776
18 -168.904
19 -173.288
20 -170.58
21 -171.624
22 -169.584
23 -165.576
24 -167.512
25 -162.416
};
\addlegendentry{k=128}
\addplot [dotted, thick, black]
table {%
1 -200
2 -199.752
3 -197.896
4 -199.356
5 -191.48
6 -194.38
7 -194.992
8 -196.916
9 -187.24
10 -196.844
11 -192.26
12 -198.036
13 -195.972
14 -197.368
15 -197.804
16 -199.752
17 -191.012
18 -194.308
19 -196.628
20 -195.424
21 -198.732
22 -199.156
23 -197.436
24 -197.924
25 -195.036
};
\addlegendentry{k=64}
\addplot [thick, green01270]
table {%
1 -110
2 -110
3 -110
4 -110
5 -110
6 -110
7 -110
8 -110
9 -110
10 -110
11 -110
12 -110
13 -110
14 -110
15 -110
16 -110
17 -110
18 -110
19 -110
20 -110
21 -110
22 -110
23 -110
24 -110
25 -110
};
\addlegendentry{sb3-A2C}
\end{axis}

\end{tikzpicture}
    \caption{Mountain Car}
    \label{fig:mountain_car}
  \end{subfigure}
  \hfill
  \begin{subfigure}{0.45\textwidth}
    \centering
\begin{tikzpicture}

\definecolor{darkgray176}{RGB}{176,176,176}
\definecolor{darkorange25512714}{RGB}{255,127,14}
\definecolor{forestgreen4416044}{RGB}{44,160,44}
\definecolor{green01270}{RGB}{0,127,0}
\definecolor{lightgray204}{RGB}{204,204,204}
\definecolor{steelblue31119180}{RGB}{31,119,180}

\begin{axis}[
legend cell align={left},
legend style={
  fill opacity=0.8,
  draw opacity=1,
  text opacity=1,
  at={(0.97,0.03)},
  anchor=south east,
  draw=lightgray204
},
tick align=outside,
tick pos=left,
x grid style={darkgray176},
xlabel={Fine-tuning Iteration},
xmin=0.3, xmax=15.7,
xtick style={color=black},
y grid style={darkgray176},
ylabel={Mean Reward},
ymin=-15.1246287025235, ymax=234.477296782843,
ytick style={color=black},
ytick={0,100,200},
yticklabels={0,100,\textcolor{white}{$-$}200},
xlabel style={yshift=0.5ex},
ylabel style={yshift=-0.5ex},
height=.22\textheight,
width=\columnwidth,
]
\path [fill=red, fill opacity=0.2]
(axis cs:1,46.5118365507413)
--(axis cs:1,12.1841634492587)
--(axis cs:2,31.1995030609939)
--(axis cs:3,17.3279571466358)
--(axis cs:4,25.3223243632786)
--(axis cs:5,29.9457090472551)
--(axis cs:6,59.8716286122037)
--(axis cs:7,100.912690545278)
--(axis cs:8,79.8388788988171)
--(axis cs:9,102.232611965941)
--(axis cs:10,121.416972557401)
--(axis cs:11,167.392068044626)
--(axis cs:12,119.668304864884)
--(axis cs:13,125.369028357119)
--(axis cs:14,176.179632759657)
--(axis cs:15,154.017086592502)
--(axis cs:15,217.622913407498)
--(axis cs:15,217.622913407498)
--(axis cs:14,213.860367240343)
--(axis cs:13,218.982971642881)
--(axis cs:12,215.611695135116)
--(axis cs:11,218.559931955374)
--(axis cs:10,223.359027442599)
--(axis cs:9,215.799388034059)
--(axis cs:8,190.137121101183)
--(axis cs:7,215.271309454722)
--(axis cs:6,184.688371387796)
--(axis cs:5,154.174290952745)
--(axis cs:4,104.269675636721)
--(axis cs:3,99.5600428533642)
--(axis cs:2,110.584496939006)
--(axis cs:1,46.5118365507413)
--cycle;

\addplot [thick, red]
table {%
1 29.348
2 70.892
3 58.444
4 64.796
5 92.06
6 122.28
7 158.092
8 134.988
9 159.016
10 172.388
11 192.976
12 167.64
13 172.176
14 195.02
15 185.82
};
\addlegendentry{k=128}
\addplot [dashed, thick, blue]
table {%
1 45.136
2 51.392
3 73.324
4 96.936
5 59.232
6 52.824
7 57.124
8 82.836
9 72.864
10 69.164
11 72.616
12 77.332
13 90.088
14 92.016
15 95.548
};
\addlegendentry{k=64}
\addplot [dotted, thick, black]
table {%
1 56.0633333333333
2 83.76
3 77.83
4 89.6533333333333
5 79.3366666666667
6 65.2566666666667
7 75.9433333333333
8 76.77
9 74.32
10 80.6766666666667
11 85.96
12 86.43
13 96.9333333333333
14 86.4566666666667
15 92.84
};
\addlegendentry{k=32}
\addplot [thick, green01270]
table {%
1 200
2 200
3 200
4 200
5 200
6 200
7 200
8 200
9 200
10 200
11 200
12 200
13 200
14 200
15 200
};
\addlegendentry{RL Target}
\end{axis}

\end{tikzpicture}
    \caption{Cartpole}
    \label{fig:cartpole}
  \end{subfigure}
  \caption{Mean and standard deviation of all experiments over multiple fine-tuning iterations}
  \label{fig:all_experiments}
\end{figure*} 

\noindent
\textbf{Acrobot.} Acrobot is a two-link pendulum that is actuated by a single joint. Its state space consists of $6$-dimensional real-valued vectors encoding link angles and angular velocities. The goal is to swing the bottom link of the pendulum up to a target height in as few steps as possible. The task is considered to be solved, if the target can be achieved in 100 steps, shown as a  green line in Fig.~\ref{fig:acrobot}. We evaluated our method with two different dimensionality reduction techniques: using a manually created dimensionality reduction via decision trees and using LDA. 
As seen in Tab.~\ref{tab:experiment_data}, both approaches can find a model which allows to solve the task in 42\% and 4\% of cases. These results are still remarkable given that our approach computes a control policy to merely maximize the probability of solving the task, but does not necessarily optimize for maximum rewards like RL agents. If we reduce the goal from 100 steps to 130 steps, we observe that our models can compute a policy that reaches this goal in 82\% and 35\% of cases. Fig.~\ref{fig:acrobot} shows the averaged gained rewards throughout learning. The  
red line indicates the average reward gained with manual dimensionality reduction and the red-shaded area
shows the standard deviation of the gained area. 
The results for LDA-based dimensionality reduction are shown in blue. We can observe that the models with manual dimensionality reduction yield to a good policy after just 4  iterations of fine-tuning (total of 200 episodes). LDA-based dimensionality
reduction leads to policies achieving less stable, but still high rewards.

\noindent
\textbf{Lunar Lander.} Lunar lander is a classic rocket trajectory optimization problem. The task  is to land the rocket in the landing area as fast as possible. At the beginning of each episode, a random force is applied to the rocket. Its state space consists of $8$-dim. vectors. We compared our learned models for two different dim. reductions with three DRL agents trained via \textit{stable-baselines}
\footnote{\url{https://huggingface.co/sb3/ppo-LunarLander-v2}}\footnote{\url{https://huggingface.co/sb3/dqn-LunarLander-v2}}\footnote{\url{https://huggingface.co/sb3/a2c-LunarLander-v2}}.
 All \textit{stable-baselines} agents are able to land the rocket successfully. As shown in Fig.~\ref{fig:lunar_lander}, the learned models are accurate enough after only 8 fine-tuning iterations
 to allow the computation of a good policy.
 The graphs follow the same color coding as for Acrobot,
 with the exception that the DRL agent results are shown
 with markers.
 We observe a performance gap between models computed from the manually-crafted dim. reduction and the LDA-based one. As seen in Table~\ref{tab:experiment_data}, the policy constructed with manually-created dim. reduction successfully lands with a probability of 81\%, compared to 35\% of the LDA-based policy. However, the policies for both models are safe and do not crash the rocket. Through visual inspection, we found that if our policy does not land successfully, it hovers close to the landing position.  It is noteworthy that the
 policy computed with manual dim. reduction even 
 outperforms two of the DRL agents and gets close 
 to the third agent that was trained using PPO~\cite{DBLP:journals/corr/SchulmanWDRK17}.

\noindent
\textbf{Mountain Car.} Mountain car is a control problem, where an agent needs to bring a car to the top of a steep hill in less than 200 steps. Environmental states consist of 2 real values, the x-position of the car and its velocity, therefore we perform
no dimensionality reduction.
As policies computed from our learned models easily complete this task, we compare to the mean reward of the agent we used to sample demonstration trajectories\footnote{\url{https://huggingface.co/sb3/a2c-MountainCar-v0}}. This reference is shown as a  
green line in Fig.~\ref{fig:mountain_car}. The 
dashed and dotted lines depict the mean reward gained
when learning a model over $k=128$ and $k=64$ clusters, respectively. The red line and red-shaded area show
the mean and standard deviation of the reward 
gained with $256$ clusters.
In this configuration, our approach with 256 clusters computed an MDP with 363 states that leads to a slightly less performant policy than the RL agent (-136$\pm$ 28 compared to -110$\pm$19). These results indicate that our method successfully learns close-to-optimal policies, even considering that it does not necessarily optimize for rewards, but towards a reachability objective. 
We further observe that a larger value for $k$ leads to a better policy. This can be attributed to the fact that a larger number of clusters helps to differentiate states during MDP learning.


\noindent
\textbf{Cartpole.} Cartpole is a classic control problem in which an agent needs to balance a pole attached to a cart. The task is considered solved with a reward of 195~\cite{DBLP:journals/tsmc/BartoSA83}, that is, the pole is balanced for at least 195 steps. This translate 
to a reward of $195$ shown as a green line in Fig.~\ref{fig:cartpole}.
Its states are encoded as 4-dimensional real-valued vectors. As seen in Tab.~\ref{tab:experiment_data}, our approach solved the game in 15 fine-tuning iterations (750 episodes) with a 213-state MDP, achieving an average reward of 195$\pm$18. Like for Mountain Car, we show
the mean and standard deviation for one configuration ($k=128$) in red and for two lower settings of $k$,
we show the mean reward. We again see that a larger
$k$ leads to better performance.

 \noindent
\textbf{Hyperparameter selection.} To select the appropriate number of clusters $k$ in k-means, we have applied the elbow method as a starting heuristic. We observed that, for the considered environments, after 128 clusters the k-means inertia decreases. 
However, after further experimentation, 
we observed that a higher number of clusters correlates with increased performance of the model but with higher computational costs, especially in the MDP learning. Therefore, we consider the perceived complexity of the environment to select a concrete $k$. 
Since we aim for very few assumptions on the environment, we used the number of dimensions of the original state space of each environment as a complexity estimate. 

The number of demonstration trajectories was set to 2500 for all experiments. This value was selected to ensure enough data points to compute sufficiently accurate dimensionality reduction and clustering accuracy. The number is further motivated by the 
number of training timesteps of one of the DRL agents we use for comparison. The Mountain Car A2C agent was trained for $10^6$ time steps, which requires at least $5000$ episodes, as the maximum episode length is $200$. Hence, we decided on using half of that for 
the initialization phase of our approach. 
The number of fine-tuning iterations, which we set to $25$, and episodes per iteration $n_{samples} = 50$ was selected so that the total number of sampled trajectories is half of the demonstration trajectories. This setting ensures relatively low sampling compared to RL agents, while the results indicate that the fine-tuning of the model can converge to a close-to-optimal solution in a few iterations.

\IOALERGIA's $\epsilon$ parameter
was set to $0.005$ for all experiments. We experimented with other values, with minimal impact on model quality. As pointed by~\cite{DBLP:journals/ita/CarrascoO99}: "The algorithm behaved robustly with respect to the choice of parameter epsilon". 

Furthermore, we observed that our approach is robust w.r.t.\ the maximal belief size, therefore we set $b_n = 4$ for all
experiments.
 
\noindent
\textbf{Discussion.}
Our experiments show that CASTLE learns deterministic labeled MDPs
with sufficient fidelity to compute policies that solve control 
tasks in their respective environments. The approach has a sample complexity comparable to DRL. 
For each task, 
we have sampled $2500$ trajectories, plus additional $1250$ trajectories per refinement iteration. \todo{BK: Martin please check this sentence. I did not understand the original one.}
For example, let's analyze the number of time steps for 
the Mountain Car experiment. We sample the environment for
approx. $3\times10^5$ time steps to sample the initial
trajectories, plus additional $1.9\times 10^5$ steps 
for the fine-tuning, which is $=4.9 \times 10^5$ time steps in total. This is less
then the sampling performed for the A2C agent 
that we use for comparison, 
which was trained for $10^6$ time steps.
In its current form, the runtime of CASTLE is slightly higher than of DRL due to the absence of GPU-accelerated computation. 
On average, a single experiment for each of the case studies took between 60 and 90 minutes, but this runtime could be improved.

While policies computed via CASTLE solved all tasks, they often achieved lower reward than DRL agents. This was to be expected since our learned models do not consider rewards and only optimize for successfully solving the task. We leave including rewards in the learned models to future work.
\emph{However, please note that the main goal of learning a compact environmental model is not to beat the performance of advanced deep RL agents.
Rather, having a compact environmental model offers many possibilities to evaluate and to explain the decision-making of DRL agents.}



\section{Conclusion}

We proposed an automata learning approach for learning 
discrete, abstract MDP models of environments 
with continuous state space and unknown stochastic
dynamics. The learning is split into two phases, a
passive and an active phase. 
In the first phase, we learn an initial model from a given set of sampled trajectories.
To prepare the data for learning, we compute a state abstraction 
by applying dimensionality reduction, clustering, and 
labeling of states. 
In the second phase, we incrementally improve the accuracy of the learned model by
actively sampling additional trajectories and use them for learning a new model. For sampling, we compute policies that maximize the probability of solving the task 
in the learned model via probabilistic model checking.
We showcase the potential of our approach by solving popular
control problems available in OpenAI gym~\cite{DBLP:journals/corr/BrockmanCPSSTZ16}. In some instances, the computed policies of the learned model achieve even higher rewards than DRL agents although we do not explicitly optimize for rewards.

We see several promising avenues for future work.
First, we want to use CASTLE to \emph{evaluate} trained RL agents for challenging application domains. The learned models can be used to analyze the agent's decision-making in crucial states. 
Furthermore, in case we detect incorrect behavior,
we want to study how we can use the models to \emph{explain} the detected issue and to guide the retraining of an agent to \emph{repair} its policy.
Another interesting line of research would be to use the learned models as \emph{runtime monitors} to detect potentially unsafe behavior of the agent during execution. Furthermore, we want to study whether the computed policies over the learned models can be used to enforce safety during runtime (aka \emph{shielding}~\cite{DBLP:conf/aaai/AlshiekhBEKNT18}).
Finally, we see several possible algorithmic extensions. For example, we want to 
enhance the learned MDPs with rewards to improve the rewards gained by the computed policies. 

\paragraph{Acknowledgments. }
This work has been supported by the "University SAL Labs" initiative of Silicon Austria Labs (SAL) and its Austrian partner universities for applied fundamental research for electronic based systems. This work also received funding from the State Government of Styria, Austria -- Department Zukunftfonds Steiermark.

\bibliography{main}
\end{document}